\pdfoutput=1

\def\year{2022}\relax
\documentclass[letterpaper]{article} 
\usepackage{aaai22}  
\usepackage{times}  
\usepackage{helvet}  
\usepackage{courier}  
\usepackage[hyphens]{url}  
\usepackage{graphicx} 
\usepackage{natbib}  
\usepackage{caption} 
\DeclareCaptionStyle{ruled}{labelfont=normalfont,labelsep=colon,strut=off} 
\usepackage{amsmath}
\usepackage{amssymb}
\usepackage{amsthm}
\usepackage{booktabs}
\usepackage{multirow}
\usepackage{subcaption}
\usepackage{svg}

\usepackage{macros}

\title{Emotion Recognition in Conversation using Probabilistic Soft Logic}

\author {
    Eriq Augustine,\textsuperscript{\rm 1}
    Connor Pryor,\textsuperscript{\rm 1}
    Pegah Jandaghi, \textsuperscript{\rm 2}
    Alon Albalak, \textsuperscript{\rm 3}
    Charles Dickens,\textsuperscript{\rm 1}
    William Wang,\textsuperscript{\rm 3}
    Lise Getoor, \textsuperscript{\rm 1}
}
\affiliations {
    \textsuperscript{\rm 1} University of California, Santa Cruz \\
    \textsuperscript{\rm 2} University of Southern California \\
    \textsuperscript{\rm 3} University of California, Santa Barbara \\
}

\begin{document}
    \maketitle
    
    \begin{abstract}
    Creating agents that can both appropriately respond to conversations and understand complex human linguistic tendencies and social cues has been a long standing challenge in the NLP community.
A recent pillar of research revolves around emotion recognition in conversation (ERC); a sub-field of emotion recognition that focuses on conversations or dialogues that contain two or more utterances.
In this work, we explore an approach to ERC that exploits the use of neural embeddings along with complex structures in dialogues.
We implement our approach in a framework called Probabilistic Soft Logic (PSL),
a declarative templating language that uses first-order like logical rules, that when combined with data, define a particular class of graphical model.
Additionally, PSL provides functionality for the incorporation of results from neural models into PSL models.
This allows our model to take advantage of advanced neural methods, such as sentence embeddings,
and logical reasoning over the structure of a dialogue.
We compare our method with state-of-the-art purely neural ERC systems,
and see almost a 20\% improvement.
With these results, we provide an extensive qualitative and quantitative analysis over the DailyDialog conversation dataset.
    \end{abstract}
    
    \section{Introduction}
\label{sec:intro}

\begin{figure}[t]
    \centering
    
    \includegraphics[width=0.46\textwidth]{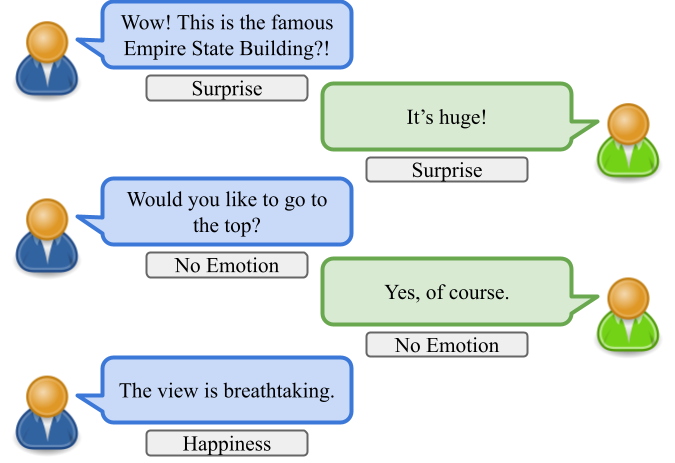}

    \caption{
        A sample conversation with representative emotions.
        Each utterance is labeled with a single emotion,
        or marked as having no emotion.
    }
    \label{fig:conversation-example}
\end{figure}

With the growing popularity of conversational agents in daily life,
the need for agents that can appropriately respond to long running conversations and that can understand complex human linguistic tendencies and social cues is becoming increasingly important.
This growth in popularity has sparked a large interest in conversational research.
A recent pillar of this emerging field has been around emotion recognition in conversation (ERC);
a sub-field of emotion recognition that focuses on conversations or dialogues that contain two or more utterances.
\citenoun{poria:access19} provides a thorough overview of the current state of ERC.
For example, in \figref{fig:conversation-example} two friends visiting the Empire State Building for the first share a typical conversation where the speakers express surprise, neutral emotion, and then happiness.
An automated assistant with access to this conversation may take very different actions depending on the emotions expressed by the speakers, e.g.,
checking for hours of operations if the emotions are positive,
or searching for other local attractions that do not involve heights if the emotions are negative.
In general, being able to correctly identify the emotion of utterances can aid other downstream tasks such as
emotion-aware dialogue agents \cite{polzin:isca00, andre:ads04, skowron:dmi10, skowron:acii11, ghandeharioun:acii19, ekbal:csiict20}
and healthcare \cite{tanana:brm11, mowafey:control12, ghandeharioun:acii19}.

ERC stands out as a challenging problem 
because it combines the already difficult task of emotion recognition
with the complexity of conversations.
Conversations are distinctly intricate because they are influenced by a variety of factors such as
topic, personality, argumentation logic, viewpoint, intent, location, number of speakers,
and the mental and emotional states of the participants at the time of the conversation
\cite{hovy:jop87, schloder:coling15, ghosal:emnlp20}.
In addition to the complexity of conversations,
ERC also have to address a number of challenges stemming from emotions, such as
bias in emotion annotations, emotional shift, and emotional reasoning \cite{poria:access19}.

In this paper, we propose a general framework that uses the structure intrinsic to dialogue
to aid in utterance emotion prediction.
Throughout this paper we develop our method using a framework called Probabilistic Soft Logic \cite{bach:jmlr17},
a declarative templating language that uses first order like logical rules, that when combined with data, define a particular class of graphical model.
PSL provides a simple framework for incorporating structural conversational knowledge through first order logical rules,
provides efficient and scalable statistical inference,
and has shown to be effective in complex domains that benefit from collective inference \cite{tomkins:lld17, kouki:kais19, sridhar:ijcai19, embar:akbc20}.
Furthermore, PSL allows for the integration of predictions from neural networks into PSL models,
allowing for the seamless use of language embeddings into structured models.

Our key contributions are as follows:
1) we create a general and extendable framework for ERC using PSL that can be applied to various ERC datasets,
2) we provide a through experimental evaluation over a popular ERC dataset, DailyDialog,
3) we show both qualitative and quantitatively that PSL outperforms the state-of-the-art models by almost 20\%, and
4) we provide a qualitative exploration of the DailyDialog dataset, in which we highlight areas of potential improvement.
    
    \section{Related Work}
\label{sec:related-work}

The broader task of emotion recognition has been a long standing problem across many fields of research, including machine learning, signal processing, social cognitive psychology, etc.
The techniques used in emotion recognition heavily overlap with the related problems of sentiment analysis and opinion mining \cite{pang:ftir08}.
All of these problems share the common goal of extracting the thoughts, feelings, and opinions of others.
However, where sentiment analysis considers a person's feelings towards an entity,
emotion recognition focuses more broadly on the emotion that a person feels, regardless of the target of that emotion.
Additionally, sentiment analysis is typically performed on more formal text sources, such as written reviews,
whereas ERC is typically performed on dialogues which are less formal and more causal in nature.

ERC has become more popular recently with the release of public conversational datasets such as social media conversations and movie/tv-show scripts \cite{zahiri:aaai18, poria:acl19}.
Recent work in ERC focuses on solving the problem with deep learning architectures.
One of the earliest networks to produce promising results for ERC was a bi-directional contextual LSTM model, bc-LSTM or CNN-cLSTM \cite{poria:acl17},
which allowed utterances to get information from subsequent or earlier utterances.
To improve upon this concept, Conversational Memory Networks (CMN) \cite{hazarika:acl18} utilizes distinct memory for each speaker to model speaker specific information.
This method was further improved by Interactive Conversational Memory Networks (ICON) \cite{hazarika:emnlp18}
and Interaction-aware Attention Networks (IAN) \cite{yeh:icassp19}, where memories were inter-connected.
DialogueRNN \cite{majumder:aaai19} expands on the previous methods by using Gated Recurrent Units (GRU) \cite{chung:neurips14} as memory cells and is specifically modeled to exploit the speaker information.
Further, DialogueGCN \cite{ghosal:emnlp19} and ConGCN \cite{zhang:ijcai19} utilize graph convolutional networks (GCN) \cite{defferrard:neurips16},
and model both context-sensitive and speaker-sensitive dependence for emotion detection.
Additionally, KET \cite{zhong:emnlp19} and COSMIC \cite{ghosal:emnlp20} attempt to improve results by using external commonsense knowledge, while BERT DCR-Net \cite{qin:aaai20} and BERT+MTL \cite{li:arxiv20} use BERT \cite{devlin:acl19} based features to aid in sentiment recognition. 
Finally, CESTa \cite{wang:sigdial20} models the ERC task as sequence tagging and uses conditional random fields (CRF) \cite{lafferty:icml01} to model the emotional consistency in conversation.

\commentout{
    In this paper, we utilize techniques from both the ERC and neural-symbolic computing (NeSy) communities.
    Neural-symbolic computing aims to incorporate logic-based reasoning with neural computation \cite{garcez:jal19}.
    This integration allows for interpretability and reasoning through symbolic knowledge,
    while providing robust learning and efficient inference with neural networks.
    \citenoun{besold:arxiv17} and \citenoun{deraedt:ijcai20} provide good surveys of recent work in this field.
}
    
    \section{Probabilistic Soft Logic}
\label{sec:psl}

Probabilistic Soft Logic (PSL) is a probabilistic programming language used to define a
special class of Markov random fields (MRF), a hinge-loss Markov random field (HL-MRF) \cite{bach:jmlr17}.
HL-MRFs are a class of conditional probabilistic models over continuous variables which allow for scalable and exact inference \cite{bach:uai13}.

PSL models relational dependencies and structural constraints using weighted first-order logical clauses,
referred to as \emph{rules}.
For example, consider the rule:

\begin{small}
    \begin{align*}
        w: & \pslpred{HasEmotion}(\pslarg{Utterance1}, \pslarg{Emotion}) \\
        & \psland \pslpred{SimilarText}(\pslarg{Utterance1}, \pslarg{Utterance2}) \\
        & \pslthen \ \pslpred{HasEmotion}(\pslarg{Utterance2}, \pslarg{Emotion})
    \end{align*}
\end{small}
where
the predicates \pslpred{HasEmotion} and \pslpred{SimilarText} respectively predict the emotional label for an utterance and define the similarity of two utterances,
and $ w $ acts as a learnable weight for the rule that denotes the rule's relative importance in the model.
This rule encodes the domain knowledge that utterances with similar texts ($\pslarg{Utterance1}$ and $\pslarg{Utterance2}$) should probably be labeled with the same emotion,
and establishes a dependency that similar utterances should share similar labels.

Given the rules for a model and data,
PSL generates an HL-MRF by instantiating concrete instances of each rule
where variables are replaced with actual entities from the data.
This process is referred to as \emph{grounding},
and each concrete instance of a rule is referred to as a \emph{ground rule}.
The logical atoms in the ground rules correspond to the random variables in the HL-MRF,
while ground rules correspond to potential functions in the HL-MRF.

Given the observed variables $ X $, unobserved variables $ Y $, and potential functions,
PSL defines a probability distribution over the unobserved variables as:
\begin{align*}
    P(Y|X) &= \frac{1}{Z(Y)} exp(-\sum_{i=1}^{m}w_i\phi_i(Y, X)) \\
    Z(Y) &= \int_{Y} exp(-\sum_{i=1}^{m}w_i\phi_i(Y, X))
\end{align*}
where $m$ is the number of potential functions,
$ \phi_i $ is the $ i^{th} $ \emph{hinge-loss potential} function,
and $w_i$ is weight of the template rule from which $ \phi_i $ was derived.
The hinge-loss potentials are defined as:
\begin{align*}
    \phi(Y, X) = [max(0, l(Y, X))]^{p}
\end{align*}
where $l$ is a linear function,
$X$ and $Y$ are in the range $[0, 1]$,
and $p \in {1, 2}$ optionally squares the potential.

Exact \textit{maximum a posteriori} (MAP) inference on this distribution can be framed as the convex optimization problem:
\begin{align*}
    Y^* & = \argmin_Y \sum_{i=1}^{m}w_i\phi_i(Y, X) \\
       & = \argmin_Y L_{map}(w, X, Y)
\end{align*}
PSL uses ADMM \cite{boyd:ftml10} to efficiently solve MAP inference.

\commentout{
    \subsection{Neural PSL}
    \label{sec:neural-psl}
    
    Neural PSL is an extension of PSL that integrates neural networks into PSL.
    This allows PSL to incorporate powerful neural techniques with PSL's logical inference.
    In Neural PSL, neural networks can be incorporated into the model as predicates.
    PSL does not restrict the number of networks that can be used in a single model,
    and places few restrictions on the types of network architectures that can be supported\footnote{
        Neural PSL accepts any neural network that can be represented in a Keras H5 file without custom code.
        This allows for the easy incorporation of networks created in the most popular deep learning frameworks, PyTorch and Tensorflow.
    }.
    This flexibility allows for multiple neural networks of different architectures to be connected in a single joint model.
    
    Neural PSL uses a method called \emph{alternating inference} to both train the neural network and solve the HL-MRF's MAP problem.
    Intuitively, Neural PSL first trains the neural networks on observed data and makes initial predictions.
    These predictions are then incorporated as observations into the the MAP problem where inference is performed in the presence of additional relational information and constraints.
    The solution to the MAP problem is then used as labels to retrain the neural networks.
    This process of alternating between training the neural network and solving the HL-MRF's MAP problem is repeated until convergence.
    
    More specifically Neural PSL defines a neural objective $ L_{nn} $ as:
    \begin{align*}
        L_{nn}(\theta, X_{nn}, X_{map}) &= \frac{1}{n} \sum^{n}_{i = 1} l(g(\theta, X_{nn}^{i}), Y_{map}^{i}) \\
        &= \frac{1}{m} \sum^{m}_{i = 1} l(Y_{nn}^{i}, Y_{map}^{i})
    \end{align*}
    where $ \theta $ are the weights of the network,
    $ g(\theta, X_{nn}^{i}) $ is the network's prediction for $ X_{nn}^{i} $ given $ \theta $,
    and $ l(Y_{nn}^{i}, Y_{map}^{i}) $ is the network's loss function, e.g., a cross-entropy loss.
    In this objective, predictions from the MAP problem, $ Y_{map} $, are used as labels.
    
    Neural PSL also redefines the PSL MAP objective, $ L_{map} $, to use predictions from the neural problem,
    $ Y_{nn} $, as additional observations:
    \begin{align*}
        L_{map}(W, X, Y) & = L_{map}(W, X_{map} \oplus Y_{nn}, Y_{map}) \\
                         & = \sum_{i=1}^{m} w_i \phi_i(Y_{map}, X_{map} \oplus Y_{nn})
    \end{align*}
    where $ X_{map} \oplus Y_{nn} $ concatenates all the observations from the MAP problem, $ X_{map} $, to predictions from the neural problem, $ Y_{nn} $.
    
    Now, Neural PSL defines a joint loss that minimizes each objective along with the distance between the
    predictions for each sub problem:
    \begin{align*}
        L_{joint} &(W, \theta, X_{map}, X_{nn}, Y_{map}, Y_{nn}) = \\
        &L_{nn}(\theta, X_{nn}, Y_{map}) \\
        &+ L_{map}(W, X_{map} \oplus Y_{nn}, Y_{map}) \\
        &+ || Y_{map} - Y_{nn} ||_2^p
    \end{align*}
    where $ p \in {1, 2} $ optionally squares the distance penalty.
    \figref{fig:neural-psl} shows the high-level interactions between the variables in Neural PSL.
    
    \begin{figure*}[tb]
        \centering
        
        \includegraphics[width=0.80 \textwidth]{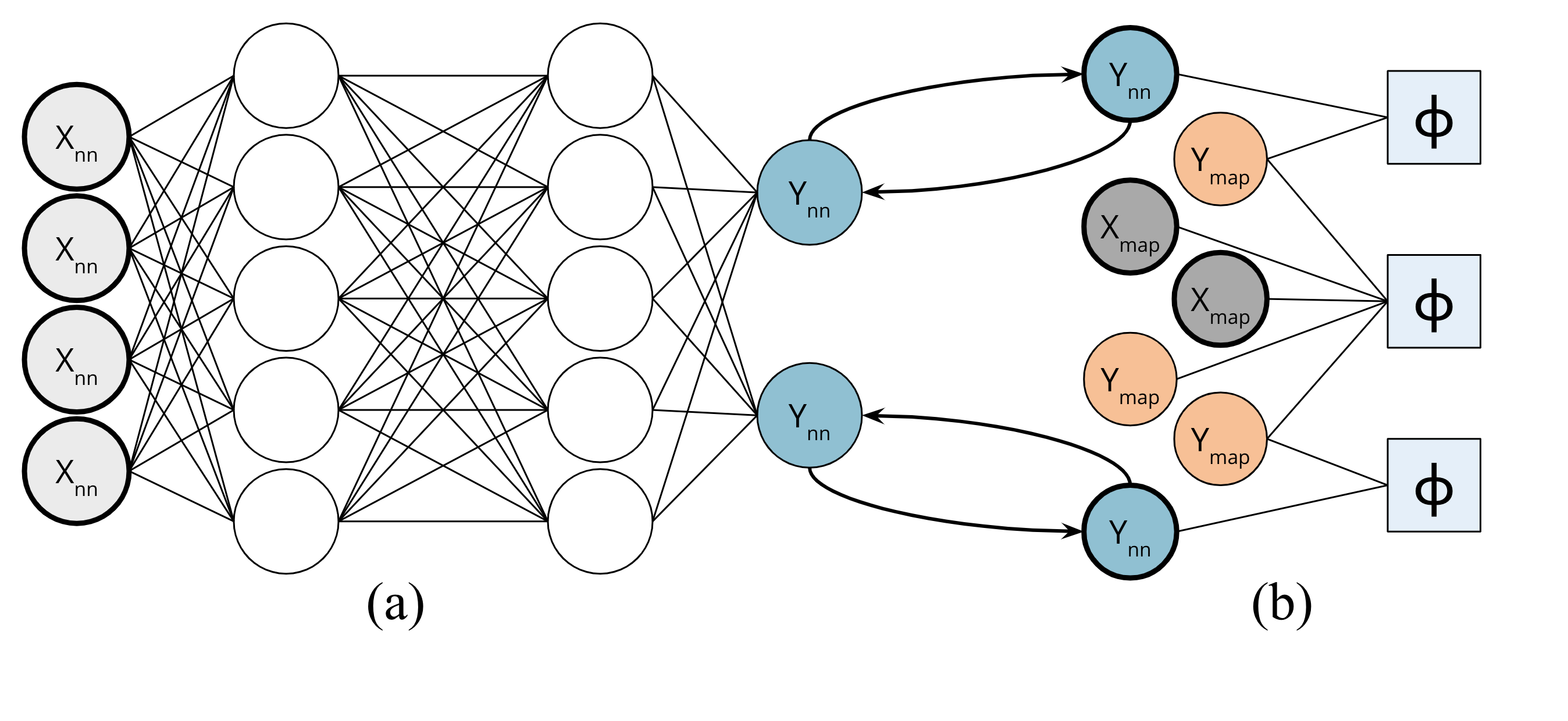}
        \vspace{-3.5mm}
    
        \caption{
            The high-level interactions of variables within Neural PSL.
            Neural PSL incorporates neural networks (a) with MAP inference over a graphical model (b).
            Predictions from the neural network $ Y_{nn} $ are used a observations in the graphical model.
            Predictions from the graphical model are then used as labels to further train the neural network.
            This process repeats until convergence.
        }
        \label{fig:neural-psl}
    \end{figure*}
}
    
    \section{ERC in PSL}
\label{sec:method}

We now describe the rules that compose our PSL model that predicts the emotion associated with each utterance.
Each rule encodes structural information about conversational emotion and can be broken into the following categories: label propagation, utterance similarity, neural classification, sum constraint, and priors.

\subsection{Label Propagation}

In this set of rules, we take advantage of the inherent structure in the dialogue to propagate labels.
First, we capture the intuition that conversations tend to have overlying dominant emotion:

\begin{small}
    \begin{align*}
        & \pslpred{NextUtterance}(\pslarg{Utterance1}, \pslarg{Utterance2}) \\ 
        & \psland \pslpred{UtteranceEmotion}(\pslarg{Utterance1}, \pslarg{Emotion}) \\
        & \pslthen \ \pslpred{UtteranceEmotion}(\pslarg{Utterance2}, \pslarg{Emotion}) \\
    \end{align*}
\end{small}%
where $ \pslpred{NextUtterance} $ ties together
an utterance, $ \pslarg{Utterance1} $ with the next utterance in the conversation $ \pslarg{Utterance2} $.
This rule propagates emotion from one utterance to the next utterance in a conversation.
In this fashion, all utterances in a conversation are chained together and an emotional shift in one influences all others.

\begin{figure}[htb]
    \centering
    
    \includegraphics[width=0.48\textwidth]{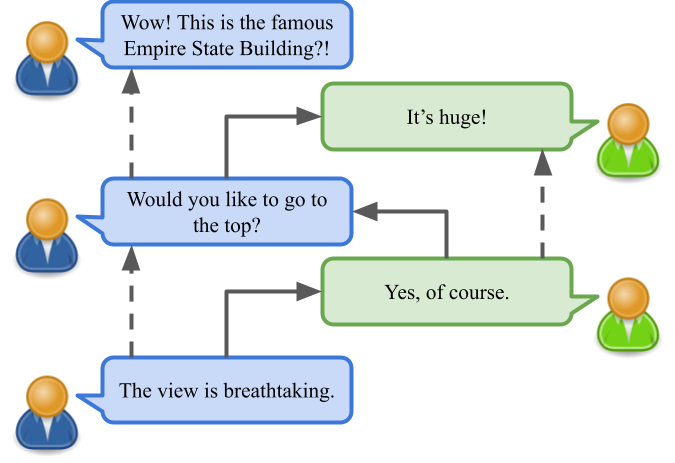}

    \caption{
        A sample conversation with the structure of the conversation displayed.
        Structure chaining together utterances with the next utterance is shown with solid arrows,
        while structure associated with a single speaker is shown with dashed arrows.
    }
    \label{fig:conversation-example-structure}
\end{figure}

The next rule models a speaker maintaining a consistent emotional state between utterances:

\begin{small}
    \begin{align*}
        & \pslpred{NextSelfUtterance}(\pslarg{Utterance1}, \pslarg{Utterance2}) \\ 
        & \psland \pslpred{UtteranceEmotion}(\pslarg{Utterance1}, \pslarg{Emotion}) \\
        & \pslthen \ \pslpred{UtteranceEmotion}(\pslarg{Utterance2}, \pslarg{Emotion}) \\
    \end{align*}
\end{small}%
where $ \pslpred{NextSelfUtterance} $ ties together
an utterance, $ \pslarg{Utterance1} $ with the next utterance spoken by the same speaker $ \pslarg{Utterance2} $.
\figref{fig:conversation-example-structure} visually demonstrates the structure captured by these two rules.

\subsection{Similarity}

This rule ensures that similar utterances have similar emotional labels:

\begin{small}
    \begin{align*}
        & \pslpred{SimilarUtterance}(\pslarg{Utterance1}, \pslarg{Utterance2}) \\ 
        & \psland \pslpred{UtteranceEmotion}(\pslarg{Utterance1}, \pslarg{Emotion}) \\
        & \pslthen \ \pslpred{UtteranceEmotion}(\pslarg{Utterance2}, \pslarg{Emotion}) \\
    \end{align*}
\end{small}%
where $ \pslpred{SimilarUtterance} $ is a computed similarity between two utterances.
Any similarity between two utterances can be used here.
In this model, we use the cosine similarity between the embeddings for each utterance.
To create embeddings, we use Google's Universal Sentence Encoder version 4 \cite{cer:acl18}.
To reduce the size of the graphical model,
we only include the highest 10 similarities for each utterance.

\subsection{Neural Classification}

\begin{figure*}[tb]
    \centering
    \includegraphics[width=0.90\textwidth]{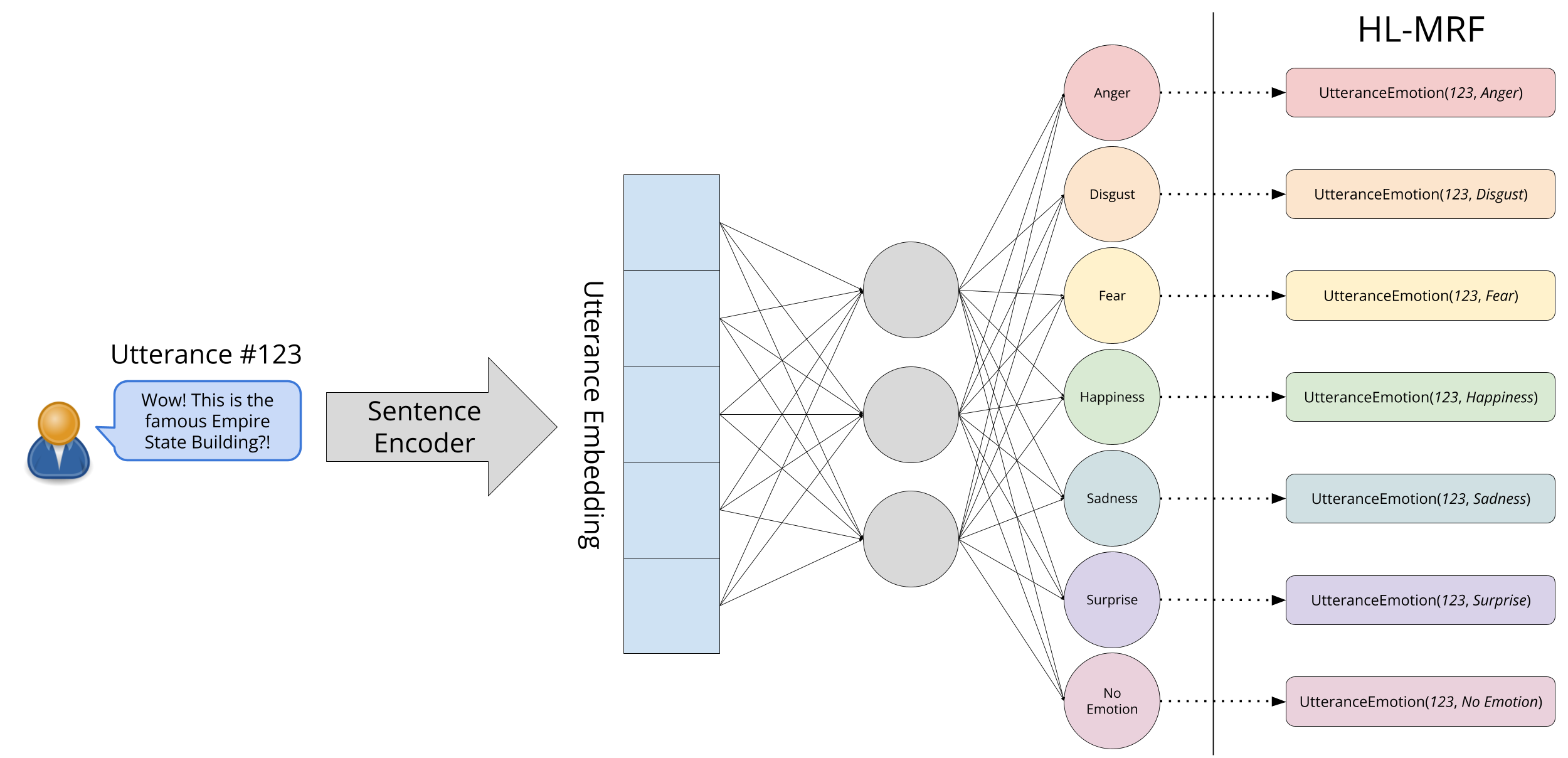}
    \caption{
        An example of how neural information is incorporated into the PSL model.
        An utterance is encoded into a sentence embedding,
        which is then passed to a neural network which makes a prediction for the emotion label.
        The predictions from the neural network are then incorporated directly into the PSL model as atoms.
    }
    \label{fig:neural-classifier}
\end{figure*}

This rule incorporates a neural model into PSL's logic-based model:

\begin{small}
    \begin{align*}
        & \pslpred{NeuralClassifier}(\pslarg{Utterance}, \pslarg{Emotion}) \\
        & \pslthen \ \pslpred{UtteranceEmotion}(\pslarg{Utterance}, \pslarg{Emotion}) \\
    \end{align*}
\end{small}%
where $ \pslpred{NeuralClassifier} $ is a neural network that takes in the embedding for an utterance,
and predicts the emotional label for that utterance.
PSL incorporates the network represented by the $ \pslpred{NeuralClassifier} $ predicate by mapping the predictions made by the network into PSL ground atoms.
\figref{fig:neural-classifier} shows how neural predictions are incorporated into the PSL model.

The network used here is a simple feedforward network with a single hidden layer.
The input is the utterance embedding,
the hidden layer has a size of 256 with a ReLu activation function,
and the output layer has one neuron per emotion and uses a softmax activation function.

\subsection{Sum Constraint}
\label{sec:sum-constraint}

Next, we use a PSL hard constraint to ensure that predictions for an utterance sum to 1:

\begin{small}
    \begin{align*}
        \pslpred{UtteranceEmotion}(\pslarg{Utterance}, \pslsum \pslarg{Emotion}) = 1.0 . \\
    \end{align*}
\end{small}%
This constraint prevents degenerate solutions where all emotions are given full or no confidence (1 and 0 respectively).
Instead all emotion predictions for an utterance must compete with one another and sum to exactly 1.

\subsection{Priors}
\label{sec:priors}

Finally, we include two negative priors into our model:

\begin{small}
    \begin{align*}
        & \pslpred{UtteranceEmotion}(\pslarg{Utterance}, \pslarg{Emotion}) = 0.0 \\
        & \pslpred{UtteranceEmotion}(\pslarg{Utterance}, '\noemotion') = 0.0 \\
    \end{align*}
\end{small}%

The first prior pushes the predictions for all utterances and emotional labels towards zero.
Pushes all values towards zero acts both as a regularizer and defaults predictions without supporting evidence to zero.

The second prior explicitly encodes the modeling assumption that every utterance is associated with an emotion.
Specifically, this rules provides an additional penalty for predicting a label of \noemotion.
This is a strong assumption that does not apply to all of ERC,
but \secref{sec:qualitative-evaluation} goes into detail on why this assumption works well with the specific dataset we used.
Therefore, in this model we assume that every utterance is associated with an emotion,
and we treat every instance of an utterance labeled without emotion as a latent variable.

In combination with the sum constraint from \secref{sec:sum-constraint},
the negative prior on \noemotion{} allow PSL to redistribute predictive mass that would otherwise be used on \noemotion{} to other class labels.
This allows our model to reason about other emotions even in the presence of a highly biased dataset like DailyDialog.

    
    \section{Dataset}
\label{sec:dataset}

The method we propose in this paper is designed to detect emotions in multi-turn dyadic conversations.
We assume that the emotional tone is fairly consistent between utterances
(i.e. there are no sudden shifts between unrelated emotions)
and emotions can propagate from one utterance to another.
These assumptions work best in conversations that are short and single topic,
such as the dialogues in \dailydialog{}.

\begin{table}[htb]
    \centering

    \begin{tabular}{ lr }
        \toprule
            Total Conversations              & 13,118 \\
            Mean Utterances Per Conversation & 7.9 \\
            Mean Tokens Per Conversation     & 114.7 \\
            Mean Tokens Per Utterance        & 14.6 \\
        \bottomrule
    \end{tabular}

    \caption{Conversation-level statistics about \dailydialog.}

    \label{table:dailydialog-conversation-stats}
\end{table}

\begin{table}[htb]
    \centering

    \begin{tabular}{ lrr }
        \toprule
            Emotion Label            & Count & Percentage \\
        \midrule
            \emotionlabel{Anger}     &  1022 &  0.99 \\
            \emotionlabel{Disgust}   &   353 &  0.34 \\
            \emotionlabel{Fear}      &    74 &  0.17 \\
            \emotionlabel{Happiness} & 12885 & 12.51 \\
            \emotionlabel{Sadness}   &  1150 &  1.12 \\
            \emotionlabel{Surprise}  &  1823 &  1.77 \\
            \noemotion               & 85572 & 83.10 \\
        \bottomrule
    \end{tabular}

    \caption{
        Label-level statistics about \dailydialog{}.
        \emph{Count} represents the total number of utterances with that emotional label (one label per utterance),
        while \emph{Percentage} represents the percentage of utterances in the dataset with the associated label.
    }

    \label{table:dailydialog-label-stats}
\end{table}

\dailydialog{} \cite{li:ijcnlp17} is a multi-turn, dyadic text dataset that was created from conversations prepared by humans for the purpose of teaching English as a second language (ESL).
Accordingly, conversations in \dailydialog{} tend to use simple vocabulary and grammatical structures.
Each conversation is designed to be a two-person conversation one may have in their typical daily communication. 
Each conversation in \dailydialog{} is short and about revolves around a specific topic.
Therefore the participants emotions in the conversations are consistent and the emotional structure of the dialogues are not complex compared to the conversations from other datasets \cite{zahiri:aaai18,poria:acl19},
which contain both long utterances and conversations are may contain about multiple topics per conversation.

The conversations in \dailydialog{} average around eight utterances split between two speakers
and cover various topics such as the weather, work life, family life, and traveling.
The \dailydialog{} dataset is partitioned into a single train-test split.
\tabref{table:dailydialog-conversation-stats} shows conversation-level statistics on this dataset.
Each utterance is labeled with one of seven emotional labels. 
The labeling for this dataset is heavily biased towards the \noemotion~label,
and to a lesser extent the \emotionlabel{Happiness}~label.
\tabref{table:dailydialog-label-stats} shows per-label statistics on this dataset.
The per-utterance emotion labels provided in \dailydialog{} allows us to incorporate the emotional structure of the dialogue during emotion detection,
which is not viable for datasets with only conversation level labels,
such as the EmpatheticDialogues dataset \cite{rashkin:arxiv18}.
    
    \section{Evaluation}
\label{sec:evaluation}

\begin{table*}[tb]
    \centering

    \begin{tabular}{ lll }
        \toprule
            Label & Prediction & Utterance \\
        \midrule
            \emotionlabel{Anger} & \emotionlabel{Disgust} &
                Yuck! \\

            \emotionlabel{Disgust} & \emotionlabel{Anger} &
                My husband goes out drinking with his friends every night. I'm fed up with it. \\
      
            \emotionlabel{Fear} & \emotionlabel{Happiness} &
                What a thrilling trip! \\
                
            \emotionlabel{Fear} & \emotionlabel{Happiness} &
                I love that dish as well. It is coconut chicken with rice. \\
                
            \emotionlabel{Fear} & \emotionlabel{Happiness} &
                I am happy that you like the house. We should write down what we like so that we can remember it. \\
                
            \emotionlabel{Happiness} & \emotionlabel{Anger} &
                Ugh! \\
                
            \emotionlabel{Surprise} & \emotionlabel{Sadness} &
                Was I? Sorry, I didn'y mean to be. I do apologize. \\

        \midrule
        
            \noemotion~& \emotionlabel{Anger} &
                Damp it!
                How are you killing me with a single shot?
                It's not fair!
                I don't want to play anymore! \\
            \noemotion~& \emotionlabel{Disgust} &
                What a creep!
                Phony good luck e-mails are one thing,
                but sexual harassment is crossing the line. \\
            \noemotion~& \emotionlabel{Fear} &
                Oh, doctor.
                Do I have to?
                I am afraid of needles! \\
            \noemotion~& \emotionlabel{Sadness} &
                I don't know, but I feel terrible. \\
            \noemotion~& \emotionlabel{Happiness} &
                And now we have a two-year-old boy.
                We're very happy that he's healthy and smart. \\
            \noemotion~& \emotionlabel{Surprise} &
                Ah! You're bleeding all over! What happened? \\

        \bottomrule
    \end{tabular}
    
    \caption{
        Utterances with likely noisy labels along with emotion predictions made by PSL.
    }

    \label{table:utterance-bad-labels}
\end{table*}

In this section, we evaluate the quantitative performance of our model against other recent ERC methods.
We also perform a qualitative analysis over our results.
Data and code will be made available upon publishing.

\subsection{Quantitative Model Comparison}
\label{sec:quantitative-evaluation}

To evaluate the performance of our model, we compare against three recent ERC models:
CNN+cLSTM \cite{poria:acl17}, COSMIC \cite{ghosal:emnlp20}, and CESTa \cite{wang:sigdial20}.
\newline

\noindent
\textbf{CNN+cLSTM} \cite{poria:acl17}:
Uses a CNN to obtain textual features for an utterance,
then applies a context LSTM (cLSTM) over those features to learn contextual information.
\newline

\noindent
\textbf{COSMIC} \cite{ghosal:emnlp20}:
Uses different elements of commonsense such as mental states, events, and causal relations
to learn interactions between interlocutors participating in a conversation.
\newline

\noindent
\textbf{CESTa} \cite{wang:sigdial20}:
Models ERC as a sequence tagging task where a conditional random field is leveraged to learn the emotional consistency in the conversation.
Uses LSTM-based encoders that capture self and inter-speaker dependency to generate contextualized utterance representations.
Uses a multi-layer transformer encoder to capture long-range global context.\vspace{5mm}\\

Following the pattern established by the previous methods,
our evaluation is performed over the single, canonical split provided with the \dailydialog{} dataset,
and the \noemotion~label is ignored when computing the Micro F1 score. \tabref{table:single-split-results} shows the results comparing our method with the previously discussed methods.
Here we can clearly see the power of incorporating structure with neural components.
Our PSL model performs nearly 20 percentage points better than the next leading method (CESTa).

To further verify our results, we evaluated our method over ten randomly generated splits of \dailydialog{}.
To create these splits,
the dataset was shuffled and 10\% of conversations were assigned to the test set
while the remaining 90\% of conversations were assigned to the train set.
For these splits, we also evaluated CNN+cLSTM to compare against our method\footnote{
    CNN+cLSTM was chosen for this comparison because of its relatively quick runtime
    and its ease-of-use when running on a new dataset.
}.
\tabref{table:single-multi-results} shows that when averaged over ten splits, PSL and CNN+cLSTM both achieve similar performance to the single canonical split.
Our PSL method diverges by only 0.33 standard deviations while CNN+cLSTM diverges by only 1.24 standard deviations.

\begin{table}[htb]
    \centering

    \begin{tabular}{ cc }
        \toprule
            Model & Micro F1 \\
        \midrule
            CNN+cLSTM & 0.518 \\
            COSMIC    & 0.585 \\
            CESTa     & 0.631 \\
            \textbf{PSL}       & \textbf{0.813} \\
        \bottomrule
    \end{tabular}

    \caption{
        Comparison of the Micro F1 of multiple methods across the canonical \dailydialog{} split.
        When Micro F1 is computed, the \noemotion~label is removed.
    }

    \label{table:single-split-results}
\end{table}

\begin{table}[htb]
    \centering

    \begin{tabular}{ cc }
        \toprule
            Model & Micro F1 \\
        \midrule
            CNN+cLSTM & 0.549 ± 0.025 \\
            \textbf{PSL}       & \textbf{0.809 ± 0.012} \\
        \bottomrule
    \end{tabular}

    \caption{
        Comparison of the Micro F1 of multiple methods across ten random \dailydialog{} splits.
        When Micro F1 is computed, the \noemotion~label is removed.
        Standard deviation is reported along with the mean Micro F1.
    }

    \label{table:single-multi-results}
\end{table}

\subsection{Noisy Emotional Labels}
\label{sec:qualitative-evaluation}

\dailydialog{} contains more than a 100k labeled utterances.
However despite being human annotated, several of the emotional labels are noisy.
Noisy labels provides an interesting challenge for ERC systems,
since these systems must overcome both the uncertain nature of human emotions in addition to the uncertain nature of noisy labels.
We posit that collective/joint methods have the potential to perform well in these noisy settings,
because relational information can provide additional signals to overpower the noisy labels.
For example, \tabref{table:utterance-bad-labels} shows several utterances that contain questionable emotion labels,
as well as the prediction PSL assigns these utterances.
In these cases, PSL provides reasonable emotional predictions over the questionable labels.

As seen in \tabref{table:dailydialog-label-stats},
\dailydialog{} is heavily biased towards the \noemotion~class.
At first, it may seem that this class represents utterances that have no clear emotional context,
as seen in \tabref{table:utterance-true-no-emotion}.
However, the \noemotion{} label is also clearly used in cases where emotional context is apparent.
\tabref{table:utterance-bad-labels} shows examples where an utterance is labeled as \noemotion,
but it is clear that a label associated with an emotion is more appropriate.
This double use of the \noemotion{} label in both cases where no clear emotion is present and where an utterance merely has no label further increases the difficulty of using the \dailydialog{} dataset.

\begin{table}[htb]
    \centering

    \begin{tabular}{ ll }
        \toprule
            Label & Utterance \\
        \midrule
            \noemotion~& I don't care. \\
            \noemotion~& Do you want black \\
                       & or white coffee? \\
            \noemotion~& She's my grandma. \\
            \noemotion~& When's your birthday? \\
            \noemotion~& I'm a doctor. \\
            \noemotion~& I certainly have. \\
            \noemotion~& About two hours ago. \\
        \bottomrule
    \end{tabular}
    
    \caption{Utterances labeled as \noemotion{} and showing no clear emotional context.}

    \label{table:utterance-true-no-emotion}
\end{table}

Finally, there are cases where the \noemotion{} label is used for a specific utterance,
but the context of the conversation provides information on what the labeling should be.
\tabref{table:utterance-context-no-emotion} shows additional dialog context
for the last two utterances in \tabref{table:utterance-true-no-emotion}.
Both of these utterances (in bold) were labeled as \noemotion{},
and without any other context that label would make sense.
However with the full context of the dialog (the speaker being bound, gagged, and robbed),
a more appropriate label should be applied
(e.g. \emotionlabel{Anger}, \emotionlabel{Fear}, or \emotionlabel{Sadness}).

\begin{table}[htb]
    \centering

    \begin{tabular}{ ll }
        \toprule
            Speaker & Utterance \\
        \midrule
            Speaker 1 & Good evening, sir. \\
                      & I understand that you have \\
                      & been robbed. \\
            Speaker 2 & \textbf{I certainly have.} \\
            Speaker 1 & When did this happen? \\
            Speaker 2 & \textbf{About two hours ago.} \\
            Speaker 1 & Why didn't you report it before? \\
            Speaker 2 & I couldn't. I was bound \\
                      & and gagged. \\
        \bottomrule
    \end{tabular}
    
    \caption{
        A conversation that demonstrates the overuse of the \noemotion~label.
        The bold utterances were labeled as \noemotion,
        but with the context of the full conversation could have been more accurately labeled.
    }

    \label{table:utterance-context-no-emotion}
\end{table}

The presence of noisy emotion labels and use of the \noemotion{} label makes \dailydialog{} a particularly difficult dataset for ERC.
However, this difficulty provides an opportunity for collective/joint methods, such as PSL, that can incorporate contextual and domain information as well as labels into predictions.
Additionally, the presence of utterances labeled \noemotion{} reinforces the modeling assumption made in \secref{sec:priors},
which assumes that all utterances contain some traces of emotion and should not be labeled \noemotion{}.
    
    \section{Conclusions and Future Work}
\label{sec:conclusion}

In this paper, we proposed a structured method for the task of ERC
that combines a simple neural model with relational inference provided by PSL.
Our initial experiments show that even a simple neural model combined with general-purpose logical rules can outperform complex and specific state-of-the-art neural models.
Furthermore, our qualitative analysis shows our model performing well even in situations where the dataset's labels are open to question.

In our future work,
we plan to extend both the neural and logical components of our model.
On the neural side, we can utilize more complex neural models.
On the logical side, we can incorporate additional structure into our models by
computing more sophisticated utterance similarity
and integrating both conversation-level and user-level similarities.
We also want to prove the generality of our approach by testing it on 
additional ERC datasets.
Finally, we plan on addressing the issues discussed in \secref{sec:qualitative-evaluation}
by relabeling the \dailydialog{} dataset with fine-grained emotion.
    
    \bibliography{augustine-starai22}
\end{document}